\documentclass[acmsmall]{acmart}

\AtBeginDocument{%
  }

\usepackage{multirow}

\begin{document}

%%
%% The "title" command has an optional parameter,
%% allowing the author to define a "short title" to be used in page headers.
\title{Shifting from endangerment to rebirth in the Artificial Intelligence Age: An Ensemble Machine Learning Approach for Hawrami Text Classification}

%%
%% The "author" command and its associated commands are used to define
%% the authors and their affiliations.
%% Of note is the shared affiliation of the first two authors, and the
%% "authornote" and "authornotemark" commands
%% used to denote shared contribution to the research.
\author{Aram Khaksar}
%\authornote{The first author designed the research method, collected data, and performed the experiment. The second author supervised the research, advised in all stages, checked the intermediary and final results at various research phases, and edited the paper for publication.}
\email{arxksr@gmail.com}
\orcid{1234-5678-9012}
\author{Hossein Hassani}
%\authornotemark[1]
\email{hosseinh@ukh.edu.krd}
\orcid{https://orcid.org/0000-0002-8899-4016}
\affiliation{%
  \institution{University of Kurdistan Hewl\^er}
  \city{Erbil}
  \state{Kurdistan Region}
  \country{Iraq}
}

%%
%% By default, the full list of authors will be used in the page
%% headers. Often, this list is too long, and will overlap
%% other information printed in the page headers. This command allows
%% the author to define a more concise list
%% of authors' names for this purpose.
%\renewcommand{\shortauthors}{Khaksar and Hassani}
\renewcommand{\shortauthors}{} %????

%%
%% The abstract is a short summary of the work to be presented in the
%% article.
\begin{abstract}
  Hawrami, a dialect of Kurdish, is classified as an endangered language as it suffers from the scarcity of data and the gradual loss of its speakers. Natural Language Processing projects can be used to partially compensate for data availability for endangered languages/dialects through a variety of approaches, such as machine translation, language model building, and corpora development. Similarly, NLP projects such as text classification are in language documentation. Several text classification studies have been conducted for Kurdish, but they were mainly dedicated to two particular dialects: Sorani (Central Kurdish) and Kurmanji (Northern Kurdish). In this paper, we introduce various text classification models using a dataset of 6,854 articles in Hawrami labeled into 15 categories by two native speakers. We use K-nearest Neighbor (KNN), Linear Support Vector Machine (Linear SVM), Logistic Regression (LR), and Decision Tree (DT) to evaluate how well those methods perform the classification task. The results indicate that the Linear SVM achieves a 96\% of accuracy and outperforms the other approaches.
\end{abstract}

\maketitle

\section{Introduction}
Language is a medium that connects individuals \citep{mustafa1971}. The extinction of a language leads to an impairment in social cohesion while also cleansing the culture and identity of its speakers. Documentation and digitization through data collection are essential in language preservation \citep{gerstenberger2017instant}. They also contribute to preserving the culture of the speakers of that language \citep{ahmadi2019towards}. Natural Language Processing (NLP) can be employed because it assists in data collection and language modeling that can help computational work on the language, hence its digital presence \citep{anastasopoulos2020endangered, zhang2022can}.

The Kurdish language consists of four dialects \citep{nabaz1976}, each owning unique scripts \citep{hassani2017blark, amini2021central}. This diversity of dialects in Kurdish introduced many challenges to NLP projects \citep{malmasi2016subdialectal, taher2023correlation}. Additionally, the speakers of each dialect encounter challenges in understanding one another. Therefore, Kurdish dialects are considered mutually unintelligible \citep{hassanpour1992nationalism, paul2008kurdish}. Attempts to reach a consensus among all scholars about its dialect classification have failed so far \citep{Kurdish2008dialectology}. For example, different claims identify Hawrami as an independent language. MacKenzie \cite{mackenzie1966dialect} explained that while Hawrami's phonetic system is similar to the rest of the Kurdish dialects, Hawrami is an independent language. In contrast, other scholars identify Hawrami as one of the Kurdish dialects since it shares common linguistic traits with the other Kurdish dialects \citep{ahmadi2020building}. In this research, we adhere to the view of a group of Kurdish scholars who classify Hawrami as a Kurdish dialect.

Kurdish as a whole is listed as a low-resourced language \citep{haig2002kurdish, esmaili2012challenges, abdulla2015sentiment}, and Hawrami is classified as `'definitely endangered'` by UNESCO \citep{moseley2010atlas}. Languages without computational support are also deemed endangered as they suffer from additional challenges in NLP and Computational Linguistics projects due to different factors, such as data scarcity, lack of standard orthography, and absence of NLP expertise familiar with them \citep{anastasopoulos2020endangered}.  Sorani and Kurmanji, the two most widely spoken dialects in Kurdish, hold a dominant position in NLP projects for Kurdish, while Hawrami has received a much lesser attention from scholars \citep{ashti2021brief, maulud2023towards}.

Through a text classification project, we aim to collect Hawrami texts from various resources (the dataset is partially publicly available for non-commercial use under the CC BY-NC-SA 4.0
license 5 at \url{https://github.com/KurdishBLARK/HawramiClassification}). Like other projects in Kurdish processing, ours also faces some well-known challenges, such as the absence of standard orthography \citep{hassani2016automatic, ahmadi2019rule}, lack of a universally recognized standard alphabet \citep{Habibullah2006Werqul, tavadze2019spreading}, and lack of standardization of the language \citep{haig2002kurdish, amini2021central}.

The rest of this paper is organized as follows. Section two provides an overview of the Hawrami dialect. Section three reflects on the related work through a literature review. Section four presents the methodology of the research. Section five illustrates the results and discusses the outcomes. Finally, section 5 concludes the paper. 

%This research uses ensemble machine-learning algorithms such as K-nearest Neighbor (KNN), Linear Support Vector Machine (Linear SVM), Logistic Regression (LR), and Decision Tree (DT) to build multiple classification models for Hawrami. Given that this approach provides the best results for text classification \citep{malmasi2016subdialectal}. It is worth mentioning that the large number of categories will make it difficult for models to classify texts correctly \citep{ikonomakis2005text}. Therefore, by considering the richness of the collected data, we prepare a dataset with the fewest categories included. Accordingly, we employ the TF-IDF vectorizer to extract features in the texts. Finally, the dataset will be publicly available on the GitHub repository of Kurdish Blark.

\section{An overview of Hawrami Dialect}

%The main obstacle to establishing a standard language for Kurdish is the lack of a standard alphabet. This issue, in particular, indicates the difference in literature and culture \citep{sabir2014standard}. In this field of study, a deep study of the history of the Kurdish alphabet is required \citep{osman2016alphabet}; nonetheless, there is still no agreement on a standard alphabet \citep{ahmadi2019rule}. These disagreements are strongly influenced by the phonetics of Kurdish \citep{hassani2017blark}.

Hawrami is spoken by the people of Hawraman, a place that traverses the border of Iran and Iraq \citep{Rzgar_2024}. There are still no credible statistics to show the accurate number of Hawrami speakers. Table \ref{tab:population} shows various sources that provide different figures of the Hawrami population.

\begin{table}[H]
  \centering
  \caption{Estimated number of Hawrami speakers}
  \label{tab:population}
  \begin{tabular}{cc}
    \toprule
    Source &Number of Speakers\\
    \midrule
    UNESCO \citep{moseley2010atlas}&23,000\\
    Paul \cite{paul2008kurdish}& $\sim$44,000 \\
    Holmberg and Odden \cite{holmberg2008noun}&\textless{}100,000 \\
    Ethnologue \cite{Ethnologue}&10,000-1000,000 \\
  \bottomrule
\end{tabular}
\end{table}

While some well-known literary works in Kurdish are written in Hawrami, through which the dialect had a significant contribution to the Kurdish literature \citep{mustafa1971}, UNESCO listed it as `'definitely endangered'`\citep{moseley2010atlas} because it seems it is gradually losing speakers and its online resources are not growing as expected. For example, currently, Wikipedia has no Hawrami article \citep{Hassani_2023}. Neamat \cite{saber2010derivational} faced challenges in their research due to the lack of resources to work on Hawrami morphemes. Arslan \cite{arslan2017zazaki} shows concerns about Hawrami preservation.

Similarly important, the classification of Hawrami is an ongoing debate among scholars. To separate dialects from languages, social and political factors must be considered \citep{hassani2017blark}. Mustafa \cite{mustafa1971} explained that the phonetics, grammar, and dictionary of all Kurdish dialects are nearly similar, and the speakers of each dialect identify themselves as Kurds. Although MacKenzie \cite{mackenzie1966dialect} showed Hawrami’s phonetic system is close to the rest of the Kurdish dialects, introduced it as a separate language from Kurdish. On the other hand, Haig \cite{haig20183} demonstrated that all Kurdish dialects share the same vowels and consonants but have different phonetics. Additionally, Ahmadi \cite{ahmadi2020building} stated that Hawrami has common linguistic traits with other Kurdish dialects. Also, Kurdish is a phonemic language \citep{amini2021central}, and its orthography highly relies on this factor.

\begin{figure}[H]
    \centering
    \begin{minipage}{.45\linewidth}
    \includegraphics[width=\linewidth]{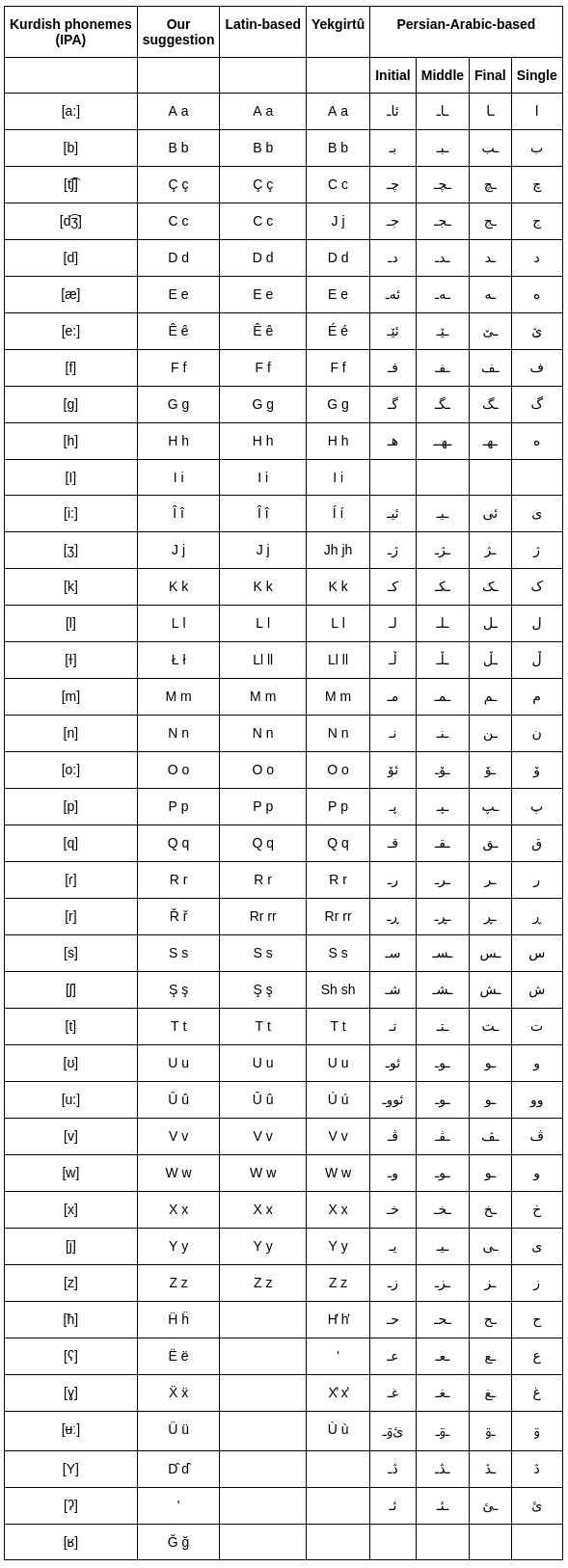}
    \end{minipage}
    \hspace{.05\linewidth}
    \begin{minipage}{.45\linewidth}
    \includegraphics[width=\linewidth]{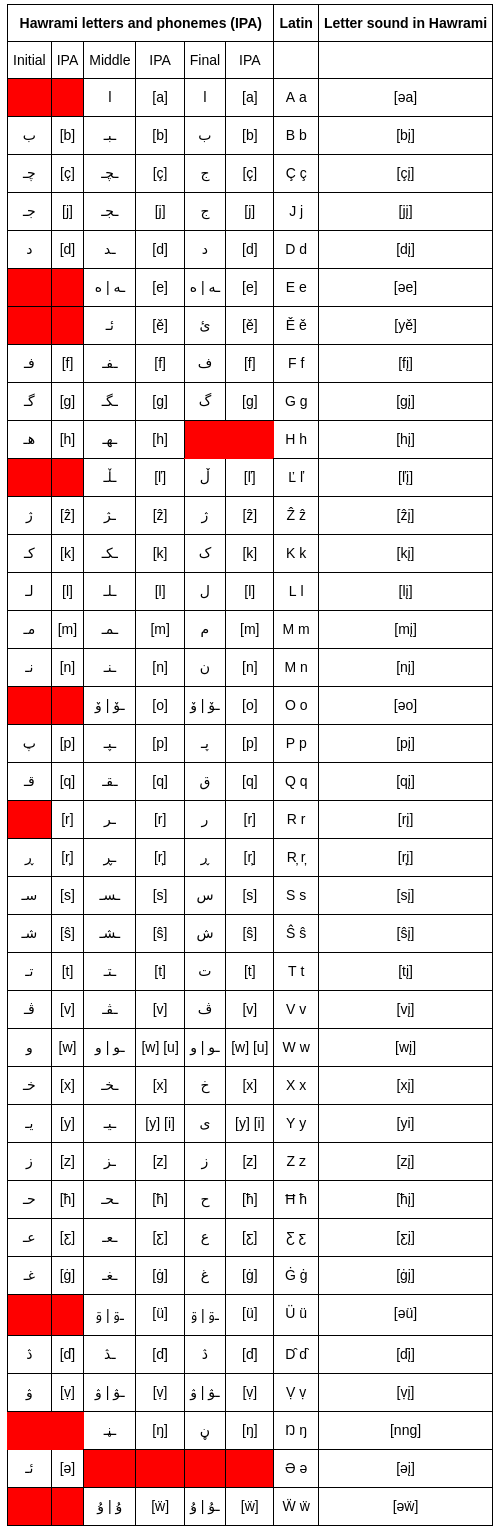}
    \end{minipage}
    \Description{Hawrami alphabets.}
    \caption{On the left: the proposed unified script Kurdish by Ahmadi et al. \cite{ahmadi2019towards}. On the right: A unified script for Hawrami is suggested by Marani \cite{Marani_2020hac}.}
    \label{fig:hac-alphabets}
\end{figure}

However, Kurdish still lacks a standard orthography \citep{ashti2021brief}. Various attempts towards standardization have not yet produced an outcome all parties, or at least a majority, have agreed upon. For example, while Ahmadi et al. \cite{ahmadi2019towards} suggested a unified script for all dialects of Kurdish, it cannot fully cover Hawrami as it lacks some phonetics that appear only in Hawrami. To address that, we introduce the Hawrami alphabets suggested by \citep{Marani_2020hac}, as shown in Figure \ref{fig:hac-alphabets}. We used this suggestion in the preparation of our dataset.

\section{Related Work}
Text classification automatically categorizes texts into different classes based on the information they contain. Text classification is a supervised machine-learning task that manually labels the text entries \citep{ikonomakis2005text, kadhim2019survey}. To find the best methods for Hawrami text classification, we found articles that provided different techniques. Accordingly, most reviews relate to research on endangered languages, as they provide reusable tools applicable to other low-resource languages.

\begin{table}[H]
 
  \caption{A summary of literature review.}
  \label{tab:lr}
  \resizebox{\textwidth}{!}{
  \begin{tabular}{ccccccc}
   \toprule
    Reference & Task & Entries & Best method(s) & Accuracy & Language\\
    \midrule
    \citep{salh2023kurdish} & Fake News Detection & 100,962 & CNN & 91.6\% & Kurdish (Sorani)\\
    \citep{9442500} & Text Classification & 60,000 & BERT & 94.48\% & Chinese\\
    \citep{abdulla2015sentiment} & Sentiment Analysis & 15,000 & Naïve Bayes & 66\% & Kurdish (Sorani)\\
    \citep{hassani2016automatic} & Text Classification & 7,000 & SVM & 92\% & Kurdish (Sorani \& Kurmanji)\\
    \citep{9016261} & Text Classification & 5,003 & SMFCNN & 95.4\% & Urdu\\
    \citep{azad2021fake} & Fake News Detection & 5,000 & SVM & 88.71\% & Kurdish (Sorani)\\
    \citep{saeed2018evaluation} & Text Classification & 4,007 & SVM & 92.4\% & Kurdish (Sorani)\\
    \citep{rashid2018automatic} & Text Classification & 4,007 & SVM & 91.03\% & Kurdish (Sorani)\\
    \citep{rashid2017systematic} & Text Classification & 2,000 & Naïve Bayes & 94.55\% & Kurdish (Sorani)\\
    \cite{sami2023sentiment}  & Sentiment Analysis & 511 & SVM \& Logistic Regression & 90\% & Kurdish (Sorani)\\
   \citep{9190447} & Text Classification & 500 & Linear SVM & 97\% & Indonesian\\
   \citep{bilianos2022greek} & Sentiment Analysis & 480 & BERT & 97\% & Greek\\
   
  \bottomrule
\end{tabular}}
\end{table}

NLP tasks can achieve excellent results if sufficient data is collected \citep{hassani2017blark}. In the previous text classification papers reviewed here, collecting data for low-resourced languages was regarded as a challenge and time-consuming.

In summary, our review found that the pipeline of a text classification project should follow: data collection, labeling, preprocessing, splitting, feature extraction, training, and testing. Furthermore, The TF-IDF method has proven to be compatible with text classification projects. Finally, using ensemble machine-learning techniques yields better results.

\section{Methodology}
\begin{figure}[]
  \centering
  \includegraphics[width=0.9\linewidth]{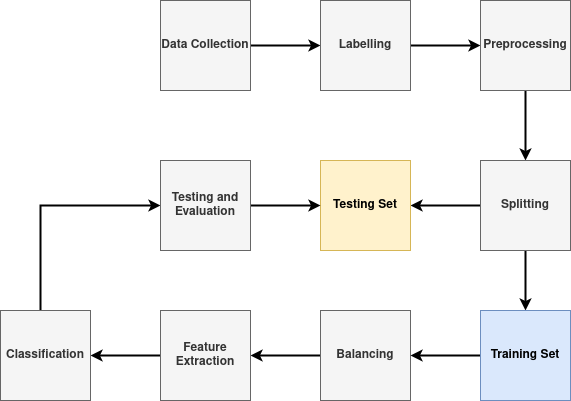}
  \caption{The proposed pipeline for text classification.}
  \Description{The methodology stages.}
  \label{fig:tpipeline}
\end{figure}

This chapter provides the processes for text classification for Hawrami, starting from collecting data, labeling, preprocessing, splitting, balancing, extracting features, classification, and testing and evaluation. The process diagram is shown in Figure \ref{fig:tpipeline}.

\subsection{Data Collection}
We employ two techniques in this project. The first approach is web scraping using the Scrapy package from Python to undertake the task. Due to Scrapy's default behavior of sending concurrent requests to the server, a process yields null data when the server is unable to respond to all the requests promptly, which results in missing data in the dataset. Scrapy also does not render dynamically handled pages. Therefore, we use Robotic Process Automation (RPA), one of the techniques used to automatically repeat human activity on browsers by creating bots to do the task with the help of the Selenium library in Python.

\subsection{Labeling}
In this stage, we provide a labeled dataset to two Hawrami speakers to review and make necessary changes. Concerning the richness of the data, there are no limitations on the number of classes. We employ various techniques to handle the classes with the fewest text entries. To measure this agreement rate, we apply Cohen's kappa, as shown in Formula \ref{eq:cohen}.
\begin{equation} \label{eq:cohen}
 k=\frac{p_{\circ}-p_{e}}{1-p_{e}}
\end{equation}

\subsection{Preprocessing}
Preprocessing is considered a crucial stage of NLP projects, in which we undertake various tasks to produce clean data for further processing in the following stage of text classification. In this stage, we accomplish the normalization task.

Normalization is the process of removing useless information from texts. Generally, the number of words in each entry differs from others, which later generates biased features. To solve this issue, we find the interquartile range for the number of words to determine outliers and calculate the 25th (first quartile) and 75th (third quartile) percentiles to specify the minimum and maximum words for each entry. For entries with a text length greater than the third quartile, we utilize the capping technique to reduce words. However, this technique produces entries with a text length below the first quartile. Therefore, we concatenate the new text entries that fall in the same category. Concatenating multiple articles with different topics from the same category does not compromise the quality of the classification model but does increase the accuracy and mitigate bias. Before capping and concatenation operations, we undertake the following tasks:
    \begin{itemize}
    \item Removing break lines, @bloggers, multiple spaces, hashtags, email addresses, punctuation marks, symbols, URLs, and numbers.
    \item Removing duplicated entries.
    \item Removing null entries.
    \item Removing Zero-width non-joiners.
    \item Removing non-Kurdish scripts.
    \item Removing stop-words.
    \item Removing words with less than two and more than fourteen characters.
    \end{itemize}
    
\subsection{Splitting}
A text classification model should identify variables capable of predicting the target variables or features and classify any text based on them. To ensure an equal distribution of entries across all classes, we split our data according to the overall distribution of the target variables. Therefore, we apply the stratified sampling technique to split the dataset into training and testing sets. This technique works well where the percentage difference between the number of entries in the major class and the mean is small. Otherwise, we apply balancing techniques to avoid reducing major classes while employing. Considering the quality of the model, we take two approaches. First, allocating only 70\% for training and the remaining 30\% for testing. In the second approach, 90\% will be taken for training and 10\% for testing.

\subsection{Balancing}
Kurdish is regarded as a low-resource language. Hence, the dataset size for Hawrami is expected to be small; the data cannot be found equally for each class. Accordingly, we calculate the percentage difference to identify the degree of imbalance. There are several techniques for balancing, including under-sampling, over-sampling, and the synthetic minority oversampling technique (SMOTE). Under-sampling trims the entries of major classes, in which data will be lost. The second technique, over-sampling, targets minor class entries and duplicates them to match the major classes. The over-sampling technique introduces a drawback as it duplicates the entire text, including useless information, and it is prone to over-fitting due to the duplication of entries. In this paper, first, we use the under-sampling technique to reduce the number of major classes and lower the percentage difference. Finally, we apply the SMOTE balancing technique to generate synthetic data by targeting features, selecting them according to the nearest neighbor from the text of minor classes, and leveraging the TF-IDF to extract those features. The balancing techniques are applied only to the training set since any resampling from the training set generates the wrong cross-validation.

\subsection{Feature Extraction}
Before feeding the texts to be trained, a feature extraction process is required to generate new sets of features from the data. For this purpose, we use Term Frequency Inverse Document Frequency (TF-IDF) to weigh each term that appears in the dataset. TF-IDF is a statistical measurement to determine the significance of a term in the entire dataset. According to the Formula \ref{eq:tf}, TF calculates the frequency of a word in a single entry, and as shown in the Formula \ref{eq:df}, DF calculates the number of articles that contain that single word. To evaluate how relevant a word is to an article, we apply the Formula \ref{eq:idf}, in which N is the total number of articles. Finally, from the Formula \ref{eq:tf-idf}, we find the weight of each word. This weight of terms indicates that the frequently appearing words are less significant. In a small dataset, we expect a small number of features, unless the vocabulary size of the collected data is large for each text entry. Considering previous facts, we identify single words as features by setting the N-gram constant to one. Additionally, all the features found in the data are used. In brief, TF-IDF reduces the high dimensionality of the data and avoids over-fitting.

\begin{equation} \label{eq:tf}
    TF_{ij} = \frac{count\ of\ i\ in\ j}{text\ length\ of\ j}
\end{equation}
\begin{equation} \label{eq:df}
    DF_i = total\ number\ of\ entries\ containing\ i
\end{equation}
\begin{equation} \label{eq:idf}
    IDF =log \frac{1 + N}{1 + DF_{i}}
\end{equation}
\begin{equation} \label{eq:tf-idf}
    TF-IDF_{ij} = TF_{ij}*IDF
\end{equation}

\subsection{Classification}
In this stage, we consider several factors according to our dataset to choose compatible classification algorithms. In a machine-learning project, the size of the dataset determines the pipeline. To avoid over-fitting, we employ simpler algorithms for smaller datasets. We train the classification model using four algorithms. The classifiers are K-Nearest Neighbor (KNN), Linear Support Vector Machine (Linear SVM), Logistic Regression (LR), and Decision Tree (DT). We build the classification models based on four scenarios:
\begin{enumerate}
        \item On the imbalanced dataset with a stratified split of 90\% for training and 10\% for testing.
        \item On the imbalanced dataset with a stratified split of 70\% for training and 30\% for testing.
        \item On the balanced dataset with a stratified split of 90\% for training and 10\% for testing.
        \item On the balanced dataset with a stratified split of 70\% for training and 30\% for testing.
\end{enumerate}

\subsubsection{K-Nearest Neighbor}
KNN is a simple yet effective supervised machine-learning algorithm for classification problems. This non-parametric algorithm is known as a lazy learner because it stores the training data for the classification and makes no assumptions about it. Known by its name, KNN is highly dependent on the value of K, which defines the number of nearest neighbors. We set different values for K to achieve the highest accuracy.

\subsubsection{Linear Support Vector Machine}
SVM is an effective supervised machine-learning algorithm employed in classification projects. This algorithm finds the hyperplanes that divide data points into different classes according to the number of features. This is achieved by determining the support vectors, the data points closest to the hyperplane. A larger distance range between the support vector and the hyperplane may improve classification; this distance is known as the margin and is measured to determine the maximum distance. In this project, we employ the linear support vector machine to divide the data points linearly.

\subsubsection{Logistic Regression}
Logistic regression is a probabilistic supervised machine-learning algorithm mostly used in classification projects. This algorithm applies the sigmoid (logistic) activation function, shown in Formula \ref{eq:sigmoid}, to find the probability that a feature belongs to a class. The logistic function expects the linear combination of a set of linearly separated features to be transformed into a probability score.
\begin{equation} \label{eq:sigmoid}
    \sigma(x)=\frac{1}{1+e^{x}}
\end{equation}

\subsubsection{Decision Tree}
A decision tree is a supervised machine-learning algorithm that consists of three main nodes: the root node, the leaf node, and the decision node. The algorithm follows a conditional approach with the help of some measures such as entropy and information gain to perform multiple tests. The process continues until it reaches a leaf node, which indicates the final classification.

\subsection{Testing and Evaluation}
We test all models on the testing set to measure the performance of each classifier using various metrics (Formulas~\ref{eq:accuracy0},~\ref{eq:accuracy1},~\ref{eq:accuracy2},~\ref{eq:accuracy3},~\ref{eq:accuracy3},~and \ref{eq:accuracy5}). 
\begin{enumerate}
    \item Accuracy: To calculate the proportion of correct predictions.
    \begin{equation} \label{eq:accuracy0}
    accuracy=\frac{correct\ predictions}{all\ predictions}=\frac{TP+TN}{TP+TN+FP+FN}
    \end{equation}
    TP: True Positive\\
    TN: True Negative\\
    FP: False Positive\\
    FN: False Negative\\
    \item Precision: To calculate the proportion of positive predictions that are actually correct.
    \begin{equation} \label{eq:accuracy1}
    precision=\frac{TP}{TP+FP}
    \end{equation}
    \item Recall: To calculate the proportion of actual positive instances that are correctly identified.
    \begin{equation} \label{eq:accuracy2}
        recall=\frac{TP}{TP+FN}
    \end{equation}
    \item F1-score: Combines precision and recall to provide a better understanding of the model's performance for imbalanced datasets.
    \begin{equation} \label{eq:accuracy3}
    F1=2\cdot\frac{precision \cdot recall}{precision+recall}
    \end{equation}
    \item F1 macro average: To calculate the model's performance equally for each class.
    \begin{equation} \label{eq:accuracy4}
    F1_{macro\ average}=\frac{sum\ of\ F1\ score\ of\ all\ classes}{total\ number\ of\ classes}=\frac{1}{C}\cdot\sum_{i=1}^{C}F1_{i}
    \end{equation}
    \item Hamming loss: To calculate the proportion of mismatched labels between the actual and predicted labels.
    \begin{equation} \label{eq:accuracy5}
    hamming\ loss=\frac{number\ of\ mismatched\ labels}{total\ number\ of\ labels}=\frac{1}{M}\cdot\sum_{i=1}^{M}d_{i}
    \end{equation}
\end{enumerate}

\section{Results}
\subsection{Dataset}
In the first stage of the experiment, we collected data from two online sources: Firat News (ANF) \citep{anf} and Kurdipedia \citep{kurdipedia}. As shown in Table \ref{tab:dataset}, we used the Scrapy framework to crawl data from the ANF website. We discovered that the dataset included empty entries and later identified the missing data type as Missing Completely at Random (MNAR). This occurred because the framework sent multiple requests per second to the server, and the website failed to respond instantly, resulting in missing data. The same scenarios occurred for Kurdipedia after using an alternate approach because their website is dynamic and Scrapy cannot render their pages. Therefore, we created a bot with the Selenium library to extract relevant texts.
\begin{table}[H]
  \caption{The data amount collected from both sources.}
  \label{tab:dataset}
  \resizebox{\textwidth}{!}{
  \begin{tabular}{ccccccc}
   \toprule
    Source & Technique & Collected & Missing & Duplicated & Deleted & After Deletion\\
    \midrule
    ANF & Web Scraping (Scrapy) & 5,317 & 44 & 3 & 1,095 & 4,222 \\
    Kurdipedia & RPA (Selenium) & 4,128 & 496 & 585 & 1,496 & 2,632 \\
    \midrule
    {Total}& & 9,445 & 61 & 68 & 2,591 & 6,854 \\
  \bottomrule
\end{tabular}}
\end{table}

\begin{figure}[H]
  \centering
  \includegraphics[width=0.9\linewidth]{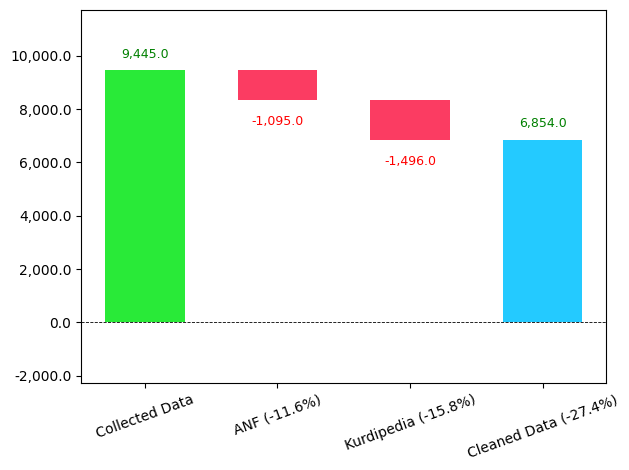}
  \caption{Total data reduction for both sources.}
  \Description{The Waterfall chart depicts the change in size of the dataset.}
  \label{fig:cleaned}
\end{figure}

As shown in Figure \ref{fig:cleaned}, 27.4\% of the data was removed, including missing data, duplicated data, non-textual data, non-Kurdish text, irrelevant content, social media content, and biased data. Finally, a dataset of 6,854 articles was created, with 65.6\% of the collected data coming from ANF and the remaining 34.4\% from Kurdipedia. 

\subsection{Labeled Data}
We prepared a labeled dataset to simplify the process for the annotators while reviewing and changing the labels as necessary. Furthermore, no limits on the number of classes were set to prevent data loss. The agreement rate found among the annotators was 93.9\%. After reviewing the dataset, we decided to use the first labeled dataset. Table \ref{tab:summary} provides an overview of all categories in the dataset. 

\begin{table}[]
  \centering
  \caption{Data summary by category.}
  \label{tab:summary}
  \resizebox{\textwidth}{!}{
  \begin{tabular}{cccccccccccccccc}
   \toprule
    \ &Politics & Literature & History & Language & Biography & Event & Women's Right & Natural Disaster& Religion & Health & Economy & Sport & Music & Cinema & Painting\\
    \midrule
    {Entries} \ & 3,749 & 2,051 & 276 & 218 & 123 & 111 & 98 & 78 & 48 & 43 & 22 & 18 & 13 & 5 & 1 \\
    {Percent} \ & 54.69 & 29.92 & 4.02 & 3.18 & 1.79 & 1.61 & 1.42 & 1.13 & 0.70 & 0.62 & 0.32 & 0.26 & 0.18 & 0.07 & 0.01 \\
    {Minimum words}   \ & 9 & 9 & 39 & 5 & 24 & 32 & 52 & 23 & 58 & 73 & 58 & 47 & 41 & 84 & 275 \\
    {Maximum words}  \ & 5,112 & 5,384 & 1,972 & 3,992 & 4,700 & 1,435 & 3,178 & 688 & 1,088 & 1,067 & 393 & 406 & 1,218 & 226 & 275 \\
    {Average words}  \ & 343.29 & 236.14 & 248.52 & 684.14 & 285.5 & 253.15 & 471.59 & 139.74 & 204.93 & 370.3 & 138.40 & 106.8 & 354.53 & 165.8 & 275 \\
    {Total words}  \ & 1,287,016 & 484,336 & 68,593 & 149,143 & 35,118 & 28,100 & 46,216 & 10,900 & 9,837 & 15,923 & 3,045 & 1,923 & 4,609 & 829 & 275 \\
    {Total unique words}   \ & 191,543 & 90,214 & 23,509 & 44,984 & 14,659 & 13,326 & 17,693 & 4,697 & 4,999 & 7,526 & 1,588 & 1,167 & 2,271 & 581 & 189 \\
    {First quartile (words)}   \ & 100 & 90 & 134 & 288.25 & 145 & 126 & 151.25 & 80 & 100.25 & 187 & 66.5 & 65.75 & 168 & 110 & 275 \\
    {Third quartile (words)}   \ & 341 & 270 & 319.5 & 780.75 & 303.5 & 304.5 & 555.75 & 172 & 202.75 & 440 & 156.75 & 123.75 & 334 & 226 & 275 \\
  \bottomrule
\end{tabular}}
\end{table}

\begin{figure}[]
  \centering
  \includegraphics[width=0.9\linewidth]{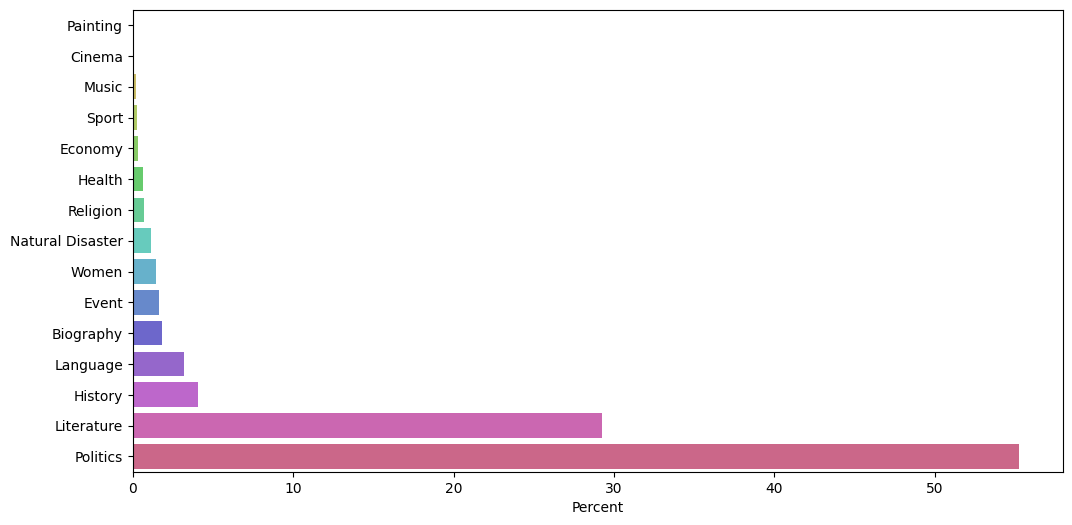}
  \caption{Distribution of data collected (percentage) across categories.}
  \Description{This chart shows the overall distribution of all categories.}
  \label{fig:classes}
\end{figure}

As shown in Figure \ref{fig:classes}, the dataset is highly biased towards the politics and literature categories, and the percentage difference is 156.92\% between the total number of political articles and the mean of the total number of articles, resulting in an imbalanced dataset.

\subsection{Preprocessed Data}
In the first step, we cleaned the texts of unnecessary information, including break lines, @bloggers, multiple spaces, hashtags, email addresses, punctuation marks, symbols, URLs, and numbers. We found that the dataset included 346 unique characters that needed to be reduced to the number of Hawrami alphabets, excluding  Perian-Arabic characters. To complete this stage, we created a dataset of all the unique characters found and assigned replacement characters or marked null. We deliberately excluded the Persian-Arabic characters, as we defined a regex particular to Hawrami to remove non-Kurdish texts. We finally reduced the number of characters to 50, and all the wrong characters were corrected according to the suggested alphabet.

A dataset of 521 Hawrami stopwords was created and used to eliminate useless words in the dataset. Furthermore, we removed double characters words. We discovered that in some of the text entries, there are meaningless texts formed from the concatenation of two or more words. To remove these texts, a method was defined to eliminate any word exceeding 14 characters. Finally, we defined a regular expression of Hawrami alphabets to identify non-Hawrami texts, which were eliminated, leaving 41 unique characters.

\begin{figure}[]
  \centering
  \includegraphics[width=0.9\linewidth]{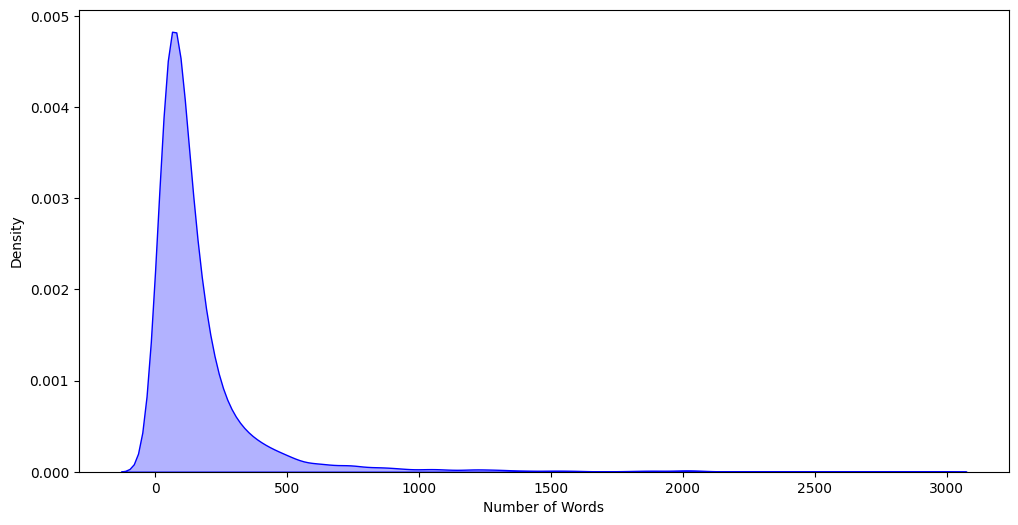}
  \caption{The overall distribution of the number of words for all entries.}
  \label{fig:skewed}
\end{figure}

\begin{figure}[]
  \centering
  \includegraphics[width=0.9\linewidth]{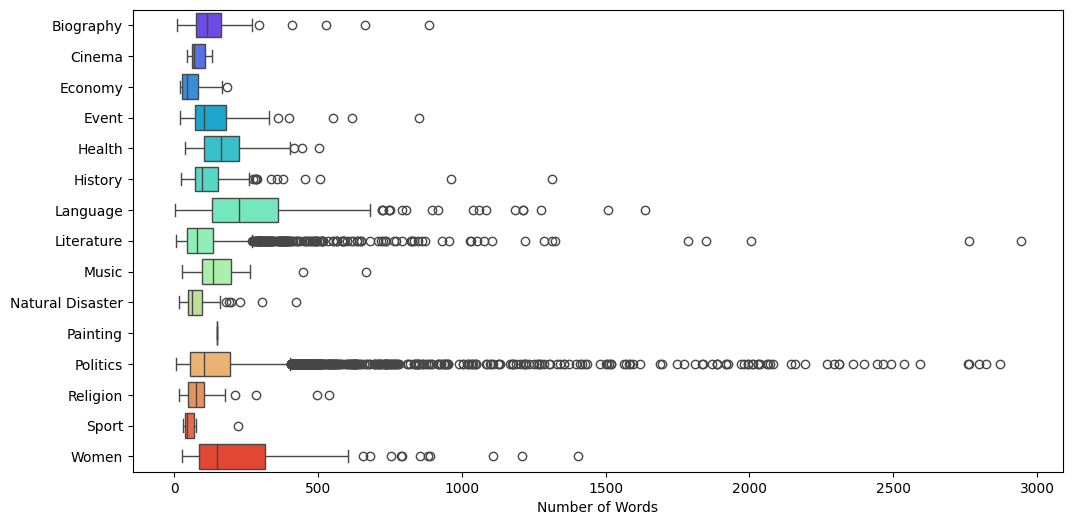}
  \caption{The range of the number of words by category.}
  \label{fig:unbalanced}
\end{figure}

As shown in Figure \ref{fig:skewed}, we observed that the number of words in each article varied and discovered a positive-skewed distribution. We also calculated the value of skewness, which was 5.34. Moreover, Figure \ref{fig:unbalanced} explicitly depicts the range of text length of entries by category in the entire dataset.

\begin{figure}[]
  \centering
  \includegraphics[width=0.9\linewidth]{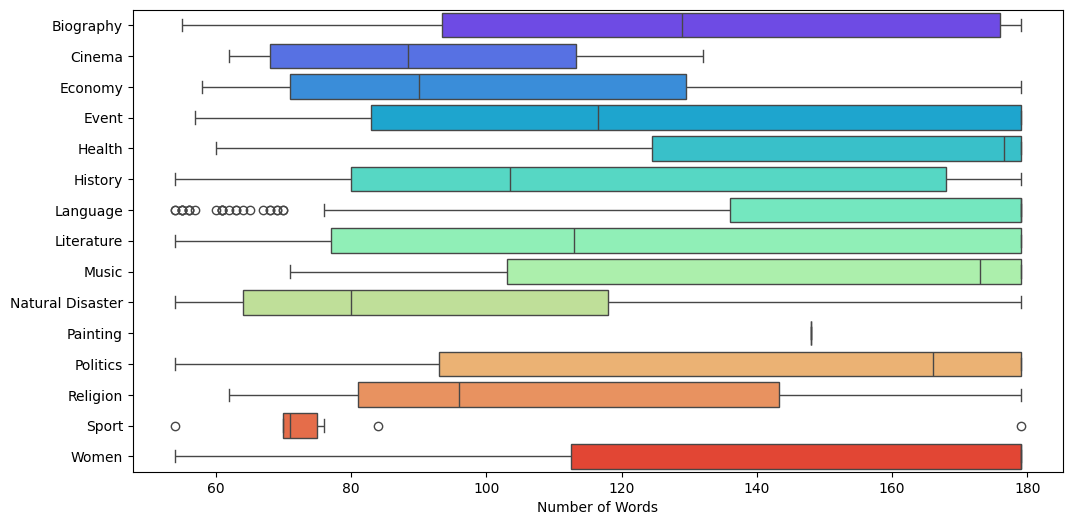}
  \caption{The range of the number of words by category after capping.}
  \label{fig:balanced}
\end{figure}

\begin{figure}[]
  \centering
  \includegraphics[width=0.9\linewidth]{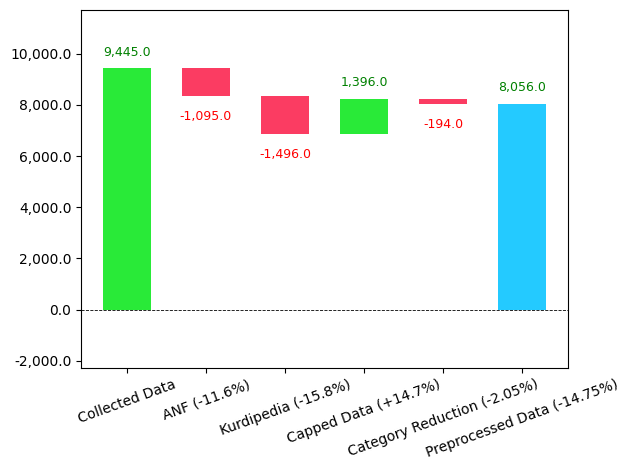}
  \caption{The dataset size after the preprocessing stage.}
  \label{fig:preprocessed}
\end{figure}

 After calculating the interquartile range, which ranged from 54 to 179 words, we detected 3,325 articles with text lengths outside this range. Therefore, we employed the capping technique to shorten the length of articles that exceeded the range. This approach produced articles with total words smaller than the first quartile. To solve this problem, we concatenated them based on their categories. As a result, the skewness ratio has fallen from 5.34 to -0.38. As shown in Figure \ref{fig:balanced}, a normal distribution for the number of words of all articles was achieved. These operations increased the number of entries from 6,854 to 8,056 as shown in Figure \ref{fig:preprocessed}.

\begin{figure}[]
  \centering
  \includegraphics[width=0.9\linewidth]{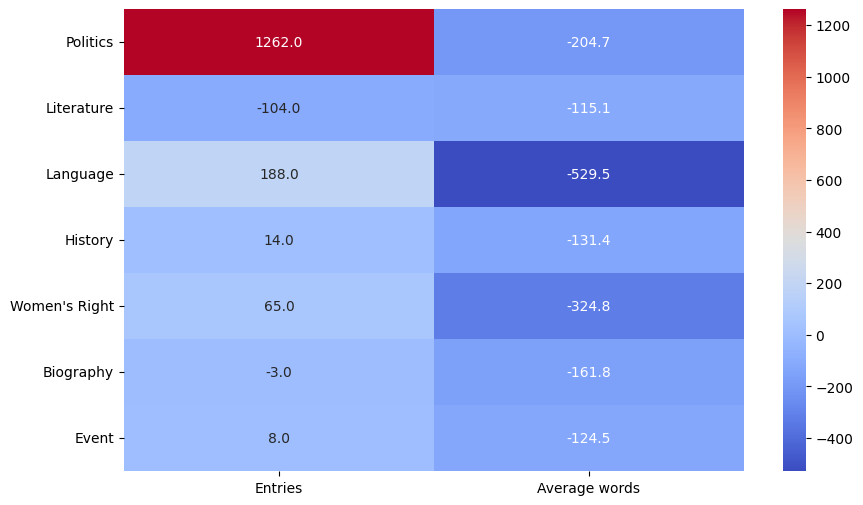}
  \caption{The difference in total entries and average words for each category after the preprocessing stage.}
  \Description{The methodology stages.}
  \label{fig:difference}
\end{figure}

In the final step, we removed articles from categories that comprise less than 1\% of the entire dataset, leaving only 7 categories as shown in Table \ref{tab:pdataset} and the percentage difference reduced to 125.66\%. As depicted in Figure \ref{fig:difference}, the preprocessing operations highly affected the dataset size, particularly for the politics category, by increasing the number of entries. Consequently, removing irrelevant words decreased the average number of words mostly of language category.

\begin{table}[H]
  \centering
  \caption{The final dataset after the preprocessing stage.}
  \label{tab:pdataset}
  \resizebox{\textwidth}{!}{
  \begin{tabular}{ccccccccccccc}
   \toprule
    \ &Politics & Literature & Language & History & Women's Right & Biography & Event \\
    \midrule
    {Entries} \ & 5,011 & 1,947 & 406 & 290 & 163 & 120 & 119  \\
    {Percent} \ & 61.2 & 24.16 & 5.03 & 3.59 & 2.02 & 1.49 & 1.47  \\
    {Minimum words}  \  & 54 & 54 & 54 & 54 & 54 & 57 & 55  \\
    {Maximum words}  \ & 179 & 179 & 179 & 179 & 179 & 179 & 179  \\
    {Average words} \  & 138.57 & 121.06 & 154.62 & 117.16 & 146.79 & 123.65 & 128.66  \\
    {Total words}  \ & 694,351 & 235,707 & 62,776 & 33,975 & 23,927 & 14,838 & 15,311  \\
    {Number of unique words}  \  & 103,168 & 87,739 & 21,489 & 13,845 & 11,299 & 7,808 & 7,879  \\
    {First quartile (words)}  \  & 93 & 77 & 136 & 80 & 112.5 & 83 & 93.5  \\
    {Third quartile (words)}  \  & 179 & 179 & 179 & 168 & 179 & 179 & 176  \\
  \bottomrule
\end{tabular}}
\end{table}

\subsection{Classification Models}
\begin{figure}[]
  \centering
  \includegraphics[width=0.9\linewidth]{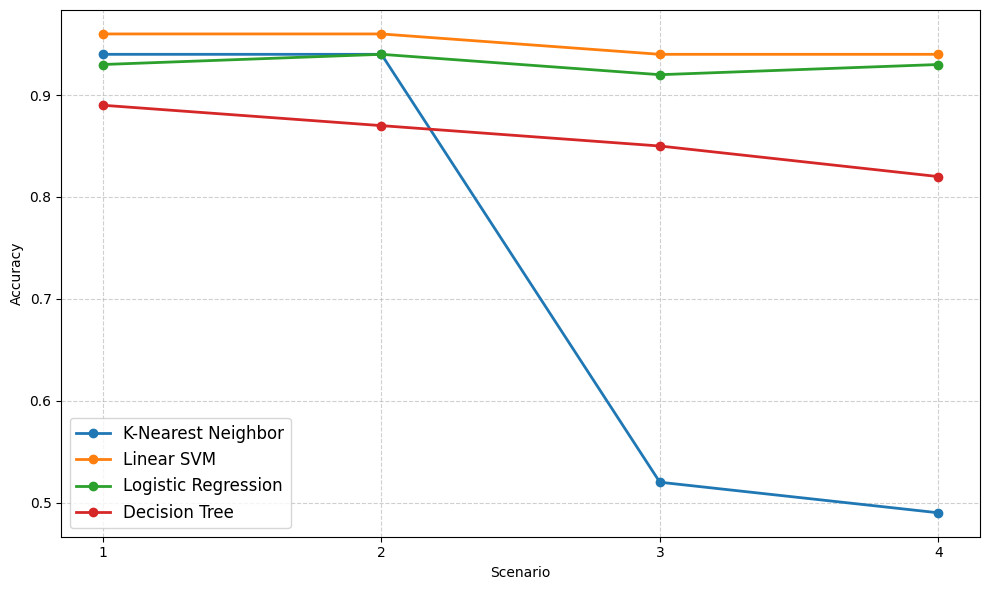}
  \caption{The performance of the models across all scenarios.}
   \label{fig:accuracy}
\end{figure}

\begin{figure}[]
  \centering
  \includegraphics[width=0.9\linewidth]{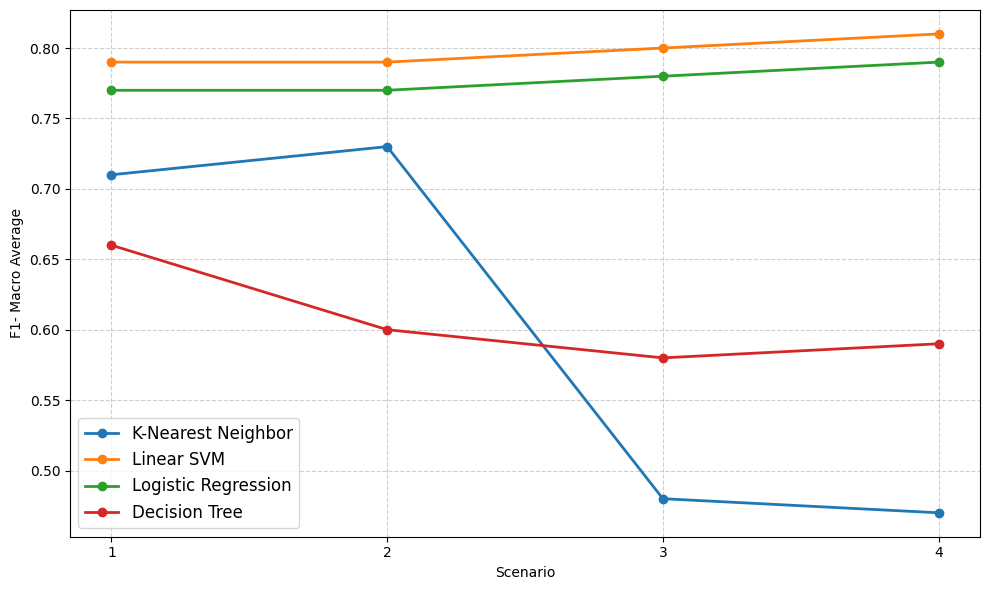}
  \caption{The balanced accuracy of the models across all scenarios.}
  \label{fig:macro}
\end{figure}

For 90\% of the training data, the TF-IDF vectorizer extracted 193,381 features and 162,546 from the 70\% training set. We applied two techniques to balance the dataset for the third and fourth scenarios. Initially, we selected only half of the political articles, comprising 31.1\% of the data, remaining with 5,551 entries. Subsequently, we applied  SMOTE to the training set, resulting in an equal number of articles for all the categories. Table \ref{tab:tests1} depicts the testing results for the classifiers across four scenarios. In Figure \ref{fig:accuracy}, the change in the accuracy of models is determined for different scenarios. Figure \ref{fig:macro} depicts the F1 macro average change of the models found in all scenarios.

\begin{table}[H]
  \centering
  \caption{The classifiers' performance across all scenarios.}
  \label{tab:tests1}
  \resizebox{\textwidth}{!}{
  \begin{tabular}{ccccccccccccc}
   \toprule
   Scenario & Algorithm & Accuracy & Precision & Recall & F1-score & F1 macro average & Hamming loss & Training time (s) \\
    \midrule
     \ & K-Nearest Neighbor & 0.94 & 0.93 & 0.94 & 0.93 & 0.71 & 0.06 & 2.52 \\
     1 & Linear SVM & 0.96 & 0.96 & 0.96 & 0.95 & 0.79 & 0.04 & 5.51 \\
     \ & Logistic Regression & 0.93 & 0.94 & 0.93 & 0.93 & 0.77 & 0.06 & 60.56 \\
     \ & Decision Tree & 0.89 & 0.90 & 0.89 & 0.90 & 0.66 & 0.1 & 14.11 \\
     \midrule
     \ & K-Nearest Neighbor & 0.94 & 0.94 & 0.94 & 0.94 & 0.73 & 0.05 & 1.51 \\
     2 & Linear SVM & 0.96 & 0.95 & 0.96 & 0.95 & 0.79 & 0.04 & 2.8 \\
     \ & Logistic Regression & 0.94 & 0.94 & 0.94 & 0.94 & 0.77 & 0.06 & 32.82 \\
     \ & Decision Tree & 0.87 & 0.88 & 0.87 & 0.87 & 0.60 & 0.13 & 10.9 \\
     \midrule
     \ & K-Nearest Neighbor & 0.52 & 0.88 & 0.52 & 0.52 & 0.48 & 0.48 & 1.4 \\
     3 & Linear SVM & 0.94 & 0.93 & 0.94 & 0.93 & 0.80 & 0.06 & 17.76 \\
     \ & Logistic Regression & 0.92 & 0.92 & 0.92 & 0.92 & 0.78 & 0.08 & 60.83 \\
     \ & Decision Tree & 0.85 & 0.84 & 0.85 & 0.84 & 0.58 & 0.15 & 25.68 \\
     \midrule
     \ & K-Nearest Neighbor & 0.49 & 0.90 & 0.49 & 0.51 & 0.47 & 0.5 & 0.09 \\
     4 & Linear SVM & 0.94 & 0.94 & 0.94 & 0.94 & 0.81 & 0.06 & 7.4 \\
     \ & Logistic Regression & 0.93 & 0.93 & 0.93 & 0.93 & 0.79 & 0.07 & 35.6 \\
     \ & Decision Tree & 0.82 & 0.83 & 0.82 & 0.82 & 0.59 & 0.18 & 16.2 \\
  \bottomrule
\end{tabular}}
\end{table}

\section{Discussion}
We employed ensemble machine-learning algorithms to build multiple text classification models for Hawrami. Accordingly, we used four algorithms, including K-Nearest Neighbor, Linear Support Vector Machine, Logistic Regression, and Decision Tree. As we obtained a small, imbalanced dataset but rich in vocabulary, we introduced four scenarios to compare the models' performance before and after balancing the dataset. Later, we applied various metrics, including accuracy, precision, recall, F1-score, F1 macro average, and hamming loss, to measure the performance of each model in different scenarios.

We under-sampled the politics category, which comprised 62.2\% of the data, and used SMOTE to create a balanced dataset. For both groups of scenarios, we created two different training and testing sets: 70\% allocated for training and 30\% for testing, and the other two scenarios used 90\% for training and only 10\% for testing. The results showed that the models' accuracy dropped for the balanced data. Considering the goal of balancing, as shown in Figure \ref{fig:macro}, the balanced accuracy is increased, implying a higher accuracy rate obtained for all the classes, particularly minor classes.

There are several reasons behind the performance of KNN on balanced data. The SMOTE technique generates synthetic data points, which increases the dimension of features in the training data. Accordingly, the KNN algorithm is prone to high-dimensional features as it fails to select the best neighbor. Therefore, more complex and robust algorithms must be trained on high-dimensional features. Due to its nature, the Decision Tree classifier poorly performed on the minor classes; the algorithm requires large data to build a deeper tree with more nodes. Even after balancing, the performance hardly improved.

\begin{figure}[]
  \centering
  \includegraphics[width=\linewidth]{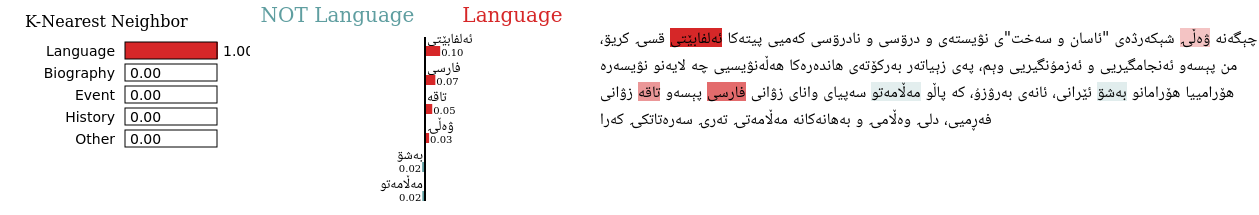}
  \includegraphics[width=\linewidth]{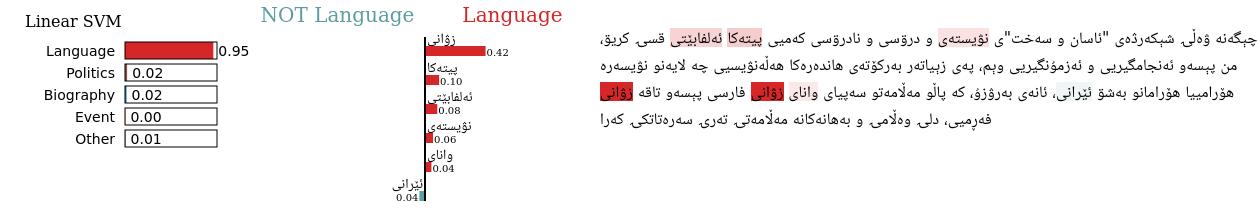}
  \includegraphics[width=\linewidth]{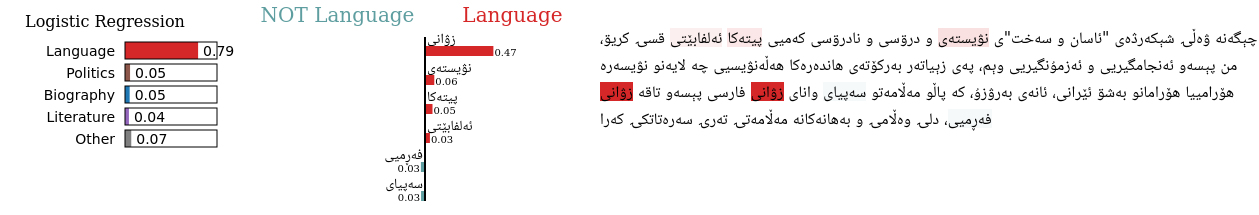}
  \includegraphics[width=\linewidth]{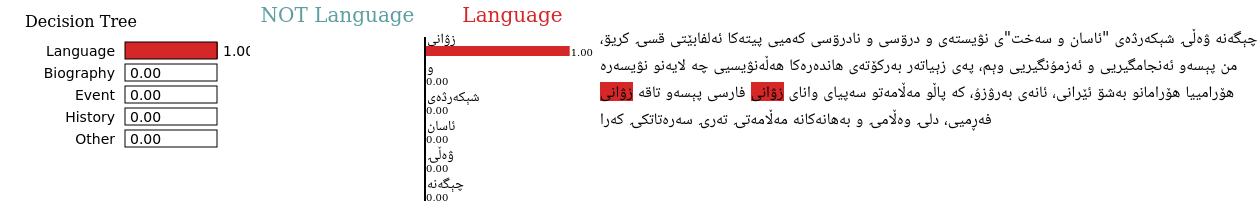}
  \caption{LIME interpretation for all the models built in the first scenario.}
  \label{fig:lime}
\end{figure}

\begin{figure}[]
  \centering
  \includegraphics[width=\linewidth]{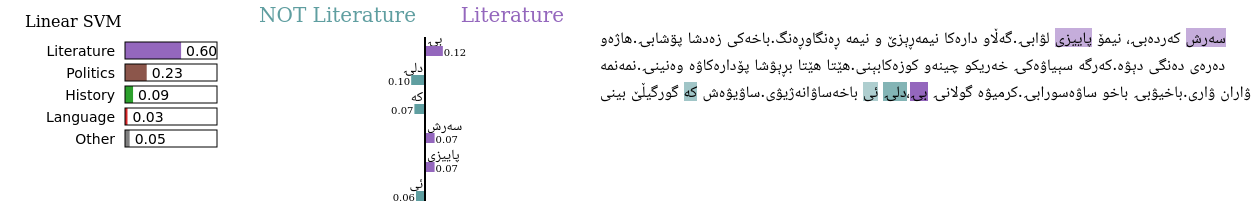}
  \includegraphics[width=\linewidth]{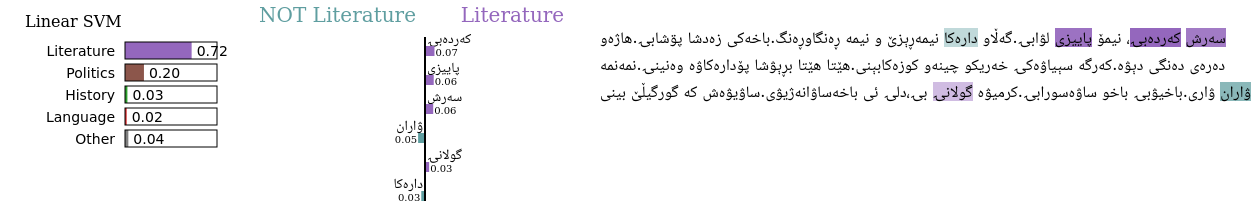}
  \caption{LIME interpretation for all the models built in the first scenario.}
  \label{fig:lime-cleaned}
\end{figure}

We provided more insights on the models built in the first scenario. Figure \ref{fig:lime} shows the prediction probabilities using the LIME interpreter to visualize the classifiers' decision-makings. All the models correctly classified the text and selected similar features to confirm their predictions. It is important to highlight that the preprocessing stage helped in creating models with great performance. Figure \ref{fig:lime-cleaned} clearly indicates that by removing irrelevant words, the model can prioritize more meaningful features to make accurate predictions on unseen data.

\begin{figure}[]
  \centering
  \includegraphics[width=\linewidth]{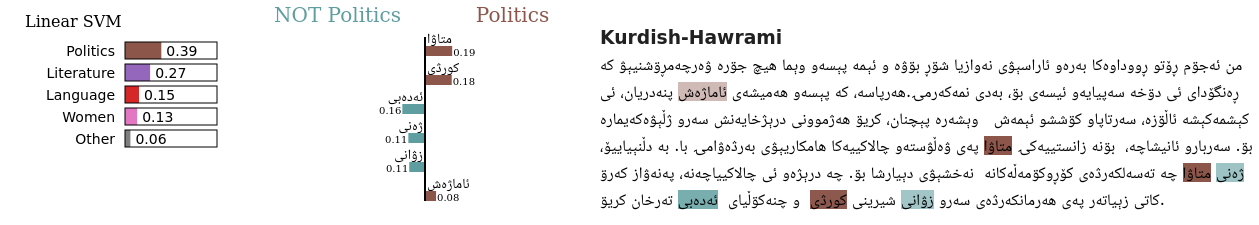}
  \includegraphics[width=\linewidth]{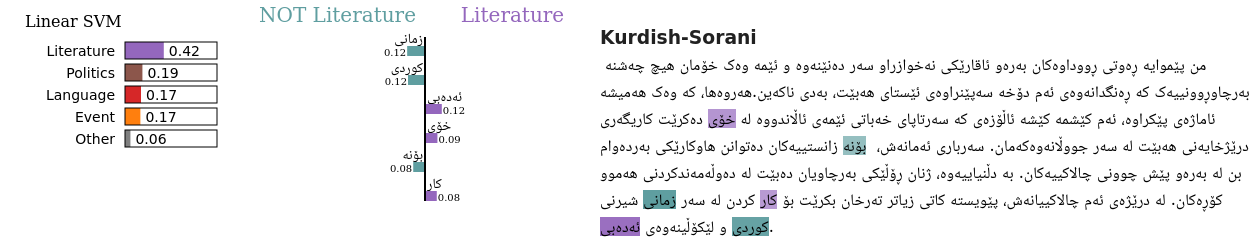}
  \caption{LIME interpretation for all the models built in the first scenario.}
  \label{fig:lime-dialects}
\end{figure}

We investigated further to evaluate the classifier's performance by generating texts that belong to multiple categories to challenge the models. We also intended to provide the translated version of the text into Sorani to observe the prediction probabilities of each classifier and identify the common traits shared between the dialects. As shown in Figure \ref{fig:lime-dialects}, the Linear SVM classifier recognized features from several categories, yet it failed to provide enough certainty. Furthermore, we noted that similar features were selected from both dialects, which supports the conclusion that Hawrami remains a branch of the Kurdish language.

Despite using a small and imbalanced dataset, the linear SVM and logistic regression scored the best accuracy in all scenarios, as shown in Table \ref{tab:tests2}. Moreover, the KNN classifier scored 94\% accuracy in the balanced dataset. A Decision Tree model scored 89\%, while not poor, as it scored the least among the best. This achievement in the performance of the models can be attributed to the richness of the data, since from 90\% of the data, the TF-IDF generated 193,381 features. This number of features in a small dataset can be exhausting for simple algorithms such as KNN, resulting in the model over-fitting. However, SVM is more robust and designed to reduce over-fitting.

\section{Conclusion and Future Works}
Hawrami is a dialect of Kurdish that is classified as an endangered language. To preserve a language, collecting textual data is considered as the first step. Additionally, these texts need to be cleaned and organized. 
    
This paper offers multiple text classification models for Hawrami using four algorithms, including K-Nearest Neighbor, Linear Support Vector Machine, Logistic Regression, and Decision Tree. We presented four scenarios to train the models on different sizes and types of data. The first two scenarios attempted to conduct experiments on the original, imbalanced dataset as well as allocate different sizes of data for training. The final scenarios were included to balance the dataset. As shown in Table \ref{tab:best-models}, the Linear SVM classifier scored the highest accuracy of 96\% in the first two scenarios.

In future works, we ask for attempts to improve techniques such as lemmatization and stemming that can be employed in the preprocessing stage to reduce the dimensionality of features. We also intend to use modern algorithms to improve the performance of the classifiers and apply feature selection methods to prevent over-fitting and select the most relevant features. Finally, we invite linguists to continue studies on the Kurdish language and share common pieces of knowledge with NLP researchers to remove the barriers that prevent Kurdish from being computationally enabled.
\begin{table}[H]
  \centering
  \caption{The best model with the highest accuracy across all scenarios.}
  \label{tab:tests2}
  \label{tab:best-models}
  \resizebox{\textwidth}{!}{
  \begin{tabular}{ccccccccccccc}
   \toprule
   Scenario & Classifier & Accuracy & Precision & Recall & F1-score & F1 macro average  \\
    \midrule
     1 & Linear SVM & 0.96 & 0.96 & 0.96 & 0.95 & 0.79  \\
     \midrule
     2 & Linear SVM & 0.96 & 0.95 & 0.96 & 0.95 & 0.79 \\
     \midrule
     3 & Linear SVM & 0.94 & 0.93 & 0.94 & 0.93 & 0.80  \\
  \midrule
     4 & Linear SVM & 0.94 & 0.94 & 0.94 & 0.94 & 0.81 \\
     
  \bottomrule
\end{tabular}}
\end{table}
% Journals use one of three template styles. All but three ACM journals
% use the {\verb|acmsmall|} template style:
% \begin{itemize}
% \item {\texttt{acmsmall}}: The default journal template style.
% \item {\texttt{acmlarge}}: Used by JOCCH and TAP.
% \item {\texttt{acmtog}}: Used by TOG.
% \end{itemize}

% \begin{description}
% \item[\texttt{sidebar}:]  Place formatted text in the margin.
% \item[\texttt{marginfigure}:] Place a figure in the margin.
% \item[\texttt{margintable}:] Place a table in the margin.
% \end{description}

%%
%% The acknowledgments section is defined using the "acks" environment
%% (and NOT an unnumbered section). This ensures the proper
%% identification of the section in the article metadata, and the
%% consistent spelling of the heading.
\begin{acks}
\begin{itemize}
    \item To Ako Marani for his great linguistic interpretations.
    \item To our annotators, Marwan Mohammed and Fardin Azimi.
    \item To Hawre Bakhawan, founder of Kurdipedia, for granting permission to use their website as a data source.
\end{itemize}

\end{acks}

%%
%% The next two lines define the bibliography style to be used, and
%% the bibliography file.
\bibliographystyle{ACM-Reference-Format}
\bibliography{HawramiClassification}

%%% -*-BibTeX-*-
%%% Do NOT edit. File created by BibTeX with style
%%% ACM-Reference-Format-Journals [18-Jan-2012].

\begin{thebibliography}{47}

%%% ====================================================================
%%% NOTE TO THE USER: you can override these defaults by providing
%%% customized versions of any of these macros before the \bibliography
%%% command.  Each of them MUST provide its own final punctuation,
%%% except for \shownote{}, \showDOI{}, and \showURL{}.  The latter two
%%% do not use final punctuation, in order to avoid confusing it with
%%% the Web address.
%%%
%%% To suppress output of a particular field, define its macro to expand
%%% to an empty string, or better, \unskip, like this:
%%%
%%% \newcommand{\showDOI}[1]{\unskip}   % LaTeX syntax
%%%
%%% \def \showDOI #1{\unskip}           % plain TeX syntax
%%%
%%% ====================================================================

\ifx \showCODEN    \undefined \def \showCODEN     #1{\unskip}     \fi
\ifx \showDOI      \undefined \def \showDOI       #1{#1}\fi
\ifx \showISBNx    \undefined \def \showISBNx     #1{\unskip}     \fi
\ifx \showISBNxiii \undefined \def \showISBNxiii  #1{\unskip}     \fi
\ifx \showISSN     \undefined \def \showISSN      #1{\unskip}     \fi
\ifx \showLCCN     \undefined \def \showLCCN      #1{\unskip}     \fi
\ifx \shownote     \undefined \def \shownote      #1{#1}          \fi
\ifx \showarticletitle \undefined \def \showarticletitle #1{#1}   \fi
\ifx \showURL      \undefined \def \showURL       {\relax}        \fi
% The following commands are used for tagged output and should be
% invisible to TeX
\providecommand\bibfield[2]{#2}
\providecommand\bibinfo[2]{#2}
\providecommand\natexlab[1]{#1}
\providecommand\showeprint[2][]{arXiv:#2}

\bibitem[anf(2024)]%
        {anf}
 \bibinfo{year}{2024}\natexlab{}.
\newblock
\newblock
\urldef\tempurl%
\url{https://anfsorani.com/%D9%87%DB%86%D8%B1%D8%A7%D9%85%DB%8C}
\showURL{%
\tempurl}


\bibitem[kur(2024)]%
        {kurdipedia}
 \bibinfo{year}{2024}\natexlab{}.
\newblock
\newblock
\urldef\tempurl%
\url{https://www.kurdipedia.org}
\showURL{%
\tempurl}


\bibitem[Abdulla and Hama(2015)]%
        {abdulla2015sentiment}
\bibfield{author}{\bibinfo{person}{Salam Abdulla} {and}
  \bibinfo{person}{Mzhda~Hiwa Hama}.} \bibinfo{year}{2015}\natexlab{}.
\newblock \showarticletitle{Sentiment analyses for Kurdish social network texts
  using Naive Bayes classifier}.
\newblock \bibinfo{journal}{\emph{Journal of University of Human Development}}
  \bibinfo{volume}{1}, \bibinfo{number}{4} (\bibinfo{year}{2015}),
  \bibinfo{pages}{393--397}.
\newblock


\bibitem[Ahmadi(2019)]%
        {ahmadi2019rule}
\bibfield{author}{\bibinfo{person}{Sina Ahmadi}.}
  \bibinfo{year}{2019}\natexlab{}.
\newblock \showarticletitle{A rule-based Kurdish text transliteration system}.
\newblock \bibinfo{journal}{\emph{ACM Transactions on Asian and Low-Resource
  Language Information Processing (TALLIP)}} \bibinfo{volume}{18},
  \bibinfo{number}{2} (\bibinfo{year}{2019}), \bibinfo{pages}{1--8}.
\newblock


\bibitem[Ahmadi(2020)]%
        {ahmadi2020building}
\bibfield{author}{\bibinfo{person}{Sina Ahmadi}.}
  \bibinfo{year}{2020}\natexlab{}.
\newblock \showarticletitle{Building a corpus for the Zaza--Gorani language
  family}. In \bibinfo{booktitle}{\emph{Proceedings of the 7th workshop on NLP
  for similar languages, varieties and dialects}}. \bibinfo{pages}{70--78}.
\newblock


\bibitem[Ahmadi et~al\mbox{.}(2019)]%
        {ahmadi2019towards}
\bibfield{author}{\bibinfo{person}{Sina Ahmadi}, \bibinfo{person}{Hossein
  Hassani}, {and} \bibinfo{person}{John~P McCrae}.}
  \bibinfo{year}{2019}\natexlab{}.
\newblock \showarticletitle{Towards electronic lexicography for the Kurdish
  language}. In \bibinfo{booktitle}{\emph{Proceedings of the sixth biennial
  conference on electronic lexicography (eLex)}}. eLex 2019.
\newblock


\bibitem[Akhter et~al\mbox{.}(2020)]%
        {9016261}
\bibfield{author}{\bibinfo{person}{Muhammad~Pervez Akhter},
  \bibinfo{person}{Zheng Jiangbin}, \bibinfo{person}{Irfan~Raza Naqvi},
  \bibinfo{person}{Mohammed Abdelmajeed}, \bibinfo{person}{Atif Mehmood}, {and}
  \bibinfo{person}{Muhammad~Tariq Sadiq}.} \bibinfo{year}{2020}\natexlab{}.
\newblock \showarticletitle{Document-Level Text Classification Using
  Single-Layer Multisize Filters Convolutional Neural Network}.
\newblock \bibinfo{journal}{\emph{IEEE Access}}  \bibinfo{volume}{8}
  (\bibinfo{year}{2020}), \bibinfo{pages}{42689--42707}.
\newblock
\urldef\tempurl%
\url{https://doi.org/10.1109/ACCESS.2020.2976744}
\showDOI{\tempurl}


\bibitem[Amalia et~al\mbox{.}(2020)]%
        {9190447}
\bibfield{author}{\bibinfo{person}{Amalia Amalia}, \bibinfo{person}{Opim~Salim
  Sitompul}, \bibinfo{person}{Erna~Budhiarti Nababan}, {and}
  \bibinfo{person}{Teddy Mantoro}.} \bibinfo{year}{2020}\natexlab{}.
\newblock \showarticletitle{An Efficient Text Classification Using fastText for
  Bahasa Indonesia Documents Classification}. In \bibinfo{booktitle}{\emph{2020
  International Conference on Data Science, Artificial Intelligence, and
  Business Analytics (DATABIA)}}. \bibinfo{pages}{69--75}.
\newblock
\urldef\tempurl%
\url{https://doi.org/10.1109/DATABIA50434.2020.9190447}
\showDOI{\tempurl}


\bibitem[Amini et~al\mbox{.}(2021)]%
        {amini2021central}
\bibfield{author}{\bibinfo{person}{Zhila Amini}, \bibinfo{person}{Mohammad
  Mohammadamini}, \bibinfo{person}{Hawre Hosseini}, \bibinfo{person}{Mehran
  Mansouri}, {and} \bibinfo{person}{Daban Jaff}.}
  \bibinfo{year}{2021}\natexlab{}.
\newblock \showarticletitle{Central Kurdish machine translation: First large
  scale parallel corpus and experiments}.
\newblock \bibinfo{journal}{\emph{arXiv preprint arXiv:2106.09325}}
  (\bibinfo{year}{2021}).
\newblock


\bibitem[Anastasopoulos et~al\mbox{.}(2020)]%
        {anastasopoulos2020endangered}
\bibfield{author}{\bibinfo{person}{Antonios Anastasopoulos},
  \bibinfo{person}{Christopher Cox}, \bibinfo{person}{Graham Neubig}, {and}
  \bibinfo{person}{Hilaria Cruz}.} \bibinfo{year}{2020}\natexlab{}.
\newblock \showarticletitle{Endangered languages meet Modern NLP}. In
  \bibinfo{booktitle}{\emph{Proceedings of the 28th International Conference on
  Computational Linguistics: Tutorial Abstracts}}. \bibinfo{pages}{39--45}.
\newblock


\bibitem[Arslan(2017)]%
        {arslan2017zazaki}
\bibfield{author}{\bibinfo{person}{Zeynep Arslan}.}
  \bibinfo{year}{2017}\natexlab{}.
\newblock \showarticletitle{Zazaki--yesterday, today and tomorrow}.
\newblock \bibinfo{journal}{\emph{Survival and standardization of a threatened
  language. Dieter Halwachs: Grazer Plurlingualismus Studien (GPS 04). Graz:
  GLM}} (\bibinfo{year}{2017}).
\newblock


\bibitem[Ashti and Kayhan(2021)]%
        {ashti2021brief}
\bibfield{author}{\bibinfo{person}{Afrasyaw Ashti} {and}
  \bibinfo{person}{Ko\c{c} Kayhan}.} \bibinfo{year}{2021}\natexlab{}.
\newblock \showarticletitle{A Brief Overview of Kurdish Natural Language
  Processing}. In \bibinfo{booktitle}{\emph{Proceedings of Workshop on
  Intelligent Information Systems}}. \bibinfo{pages}{185}.
\newblock


\bibitem[Azad et~al\mbox{.}(2021)]%
        {azad2021fake}
\bibfield{author}{\bibinfo{person}{Rania Azad}, \bibinfo{person}{Bilal
  Mohammed}, \bibinfo{person}{Rawaz Mahmud}, \bibinfo{person}{Lanya Zrar},
  {and} \bibinfo{person}{Shajwan Sdiqa}.} \bibinfo{year}{2021}\natexlab{}.
\newblock \showarticletitle{Fake News Detection in low-resourced languages
  {Kurdish language} using Machine learning algorithms}.
\newblock \bibinfo{journal}{\emph{Turkish Journal of Computer and Mathematics
  Education (TURCOMAT)}} \bibinfo{volume}{12}, \bibinfo{number}{6}
  (\bibinfo{year}{2021}), \bibinfo{pages}{4219--4225}.
\newblock


\bibitem[Bilianos(2022)]%
        {bilianos2022greek}
\bibfield{author}{\bibinfo{person}{Dimitris Bilianos}.}
  \bibinfo{year}{2022}\natexlab{}.
\newblock \showarticletitle{Experiments in Text Classification: Analyzing the
  Sentiment of Electronic Product Reviews in Greek}.
\newblock \bibinfo{journal}{\emph{Journal of Quantitative Linguistics}}
  \bibinfo{volume}{29}, \bibinfo{number}{3} (\bibinfo{year}{2022}),
  \bibinfo{pages}{374--386}.
\newblock
\urldef\tempurl%
\url{https://doi.org/10.1080/09296174.2021.1885872}
\showDOI{\tempurl}
\showeprint{https://doi.org/10.1080/09296174.2021.1885872}


\bibitem[Esmaili(2012)]%
        {esmaili2012challenges}
\bibfield{author}{\bibinfo{person}{Kyumars~Sheykh Esmaili}.}
  \bibinfo{year}{2012}\natexlab{}.
\newblock \showarticletitle{Challenges in Kurdish text processing}.
\newblock \bibinfo{journal}{\emph{arXiv preprint arXiv:1212.0074}}
  (\bibinfo{year}{2012}).
\newblock


\bibitem[Ethnologue({[n.\,d.]})]%
        {Ethnologue}
\bibfield{author}{\bibinfo{person}{Ethnologue}.}
  \bibinfo{year}{[n.\,d.]}\natexlab{}.
\newblock
\newblock
\urldef\tempurl%
\url{https://www.ethnologue.com/language/hac/}
\showURL{%
\tempurl}


\bibitem[Gerstenberger et~al\mbox{.}(2017)]%
        {gerstenberger2017instant}
\bibfield{author}{\bibinfo{person}{Ciprian Gerstenberger},
  \bibinfo{person}{Niko Partanen}, \bibinfo{person}{Michael Rie{\ss}ler}, {and}
  \bibinfo{person}{Joshua Wilbur}.} \bibinfo{year}{2017}\natexlab{}.
\newblock \showarticletitle{Instant annotations--Applying NLP methods to the
  annotation of spoken language documentation corpora}. In
  \bibinfo{booktitle}{\emph{Proceedings of the Third Workshop on Computational
  Linguistics for Uralic Languages}}. \bibinfo{pages}{25--36}.
\newblock


\bibitem[Habibullah(2006)]%
        {Habibullah2006Werqul}
\bibfield{author}{\bibinfo{person}{Jamal Habibullah}.}
  \bibinfo{year}{2006}\natexlab{}.
\newblock \bibinfo{booktitle}{\emph{Ferheng\^i Werq\^ul}}.
\newblock \bibinfo{publisher}{Sp\^ir\^ez}, \bibinfo{address}{Duhok}.
\newblock


\bibitem[Haig(2018)]%
        {haig20183}
\bibfield{author}{\bibinfo{person}{Geoffrey Haig}.}
  \bibinfo{year}{2018}\natexlab{}.
\newblock \showarticletitle{3.3. The Iranian languages of northern Iraq}.
\newblock \bibinfo{journal}{\emph{The Languages and Linguistics of Western
  Asia: An Areal Perspective}}  \bibinfo{volume}{6} (\bibinfo{year}{2018}),
  \bibinfo{pages}{267}.
\newblock


\bibitem[Haig and Matras(2002)]%
        {haig2002kurdish}
\bibfield{author}{\bibinfo{person}{Geoffrey Haig} {and} \bibinfo{person}{Yaron
  Matras}.} \bibinfo{year}{2002}\natexlab{}.
\newblock \showarticletitle{Kurdish linguistics: a brief overview}.
\newblock \bibinfo{journal}{\emph{STUF-Language Typology and Universals}}
  \bibinfo{volume}{55}, \bibinfo{number}{1} (\bibinfo{year}{2002}),
  \bibinfo{pages}{3--14}.
\newblock


\bibitem[Hassani(2017)]%
        {hassani2017blark}
\bibfield{author}{\bibinfo{person}{Hossein Hassani}.}
  \bibinfo{year}{2017}\natexlab{}.
\newblock \showarticletitle{BLARK for multi-dialect languages: towards the
  Kurdish BLARK}.
\newblock \bibinfo{journal}{\emph{Language Resources and Evaluation}}
  \bibinfo{volume}{52}, \bibinfo{number}{2} (\bibinfo{year}{2017}),
  \bibinfo{pages}{625--644}.
\newblock


\bibitem[Hassani(2023)]%
        {Hassani_2023}
\bibfield{author}{\bibinfo{person}{Hossein Hassani}.}
  \bibinfo{year}{2023}\natexlab{}.
\newblock \bibinfo{title}{The impact of the Kurdish language on Technological
  Singularity. | Hussein Hassani | TEDxNishtiman}.
\newblock
\newblock
\urldef\tempurl%
\url{https://www.youtube.com/watch?v=sDvs_8hMDzA&t=78s}
\showURL{%
\tempurl}


\bibitem[Hassani et~al\mbox{.}(2016)]%
        {hassani2016automatic}
\bibfield{author}{\bibinfo{person}{Hossein Hassani}, \bibinfo{person}{Dzejla
  Medjedovic}, {et~al\mbox{.}}} \bibinfo{year}{2016}\natexlab{}.
\newblock \showarticletitle{Automatic Kurdish dialects identification}.
\newblock \bibinfo{journal}{\emph{Computer Science \& Information Technology}}
  \bibinfo{volume}{6}, \bibinfo{number}{2} (\bibinfo{year}{2016}),
  \bibinfo{pages}{61--78}.
\newblock


\bibitem[Hassanpour(1992)]%
        {hassanpour1992nationalism}
\bibfield{author}{\bibinfo{person}{Amir Hassanpour}.}
  \bibinfo{year}{1992}\natexlab{}.
\newblock \showarticletitle{Nationalism and language in Kurdistan, 1918-1985}.
\newblock  (\bibinfo{year}{1992}).
\newblock


\bibitem[Holmberg and Odden(2008)]%
        {holmberg2008noun}
\bibfield{author}{\bibinfo{person}{Anders Holmberg} {and}
  \bibinfo{person}{David Odden}.} \bibinfo{year}{2008}\natexlab{}.
\newblock \showarticletitle{The noun phrase in Hawrami}.
\newblock \bibinfo{journal}{\emph{Aspects of Iranian linguistics}}
  (\bibinfo{year}{2008}).
\newblock


\bibitem[Ikonomakis et~al\mbox{.}(2005)]%
        {ikonomakis2005text}
\bibfield{author}{\bibinfo{person}{M Ikonomakis}, \bibinfo{person}{Sotiris
  Kotsiantis}, \bibinfo{person}{Vasilis Tampakas}, {et~al\mbox{.}}}
  \bibinfo{year}{2005}\natexlab{}.
\newblock \showarticletitle{Text classification using machine learning
  techniques.}
\newblock \bibinfo{journal}{\emph{WSEAS transactions on computers}}
  \bibinfo{volume}{4}, \bibinfo{number}{8} (\bibinfo{year}{2005}),
  \bibinfo{pages}{966--974}.
\newblock


\bibitem[Kadhim(2019)]%
        {kadhim2019survey}
\bibfield{author}{\bibinfo{person}{Ammar~Ismael Kadhim}.}
  \bibinfo{year}{2019}\natexlab{}.
\newblock \showarticletitle{Survey on supervised machine learning techniques
  for automatic text classification}.
\newblock \bibinfo{journal}{\emph{Artificial Intelligence Review}}
  \bibinfo{volume}{52}, \bibinfo{number}{1} (\bibinfo{year}{2019}),
  \bibinfo{pages}{273--292}.
\newblock


\bibitem[Kurdish Academy~of Language(2016)]%
        {Kurdish2008dialectology}
\bibfield{author}{\bibinfo{person}{KAL Kurdish Academy~of Language}.}
  \bibinfo{year}{2016}\natexlab{}.
\newblock \bibinfo{title}{Kurdish dialectology}.
\newblock
\newblock
\urldef\tempurl%
\url{https://kurdishacademy.org/?p=373}
\showURL{%
\tempurl}


\bibitem[MacKenzie(1966)]%
        {mackenzie1966dialect}
\bibfield{author}{\bibinfo{person}{David~Neil MacKenzie}.}
  \bibinfo{year}{1966}\natexlab{}.
\newblock \bibinfo{booktitle}{\emph{The dialect of Awroman
  (Hawr{\=a}m{\=a}n-{\=\i} Luh{\=o}n): Grammatical sketch, texts, and
  vocabulary}}.
\newblock \bibinfo{publisher}{Munksgaard, Elisabeth}.
\newblock


\bibitem[Malmasi(2016)]%
        {malmasi2016subdialectal}
\bibfield{author}{\bibinfo{person}{Shervin Malmasi}.}
  \bibinfo{year}{2016}\natexlab{}.
\newblock \showarticletitle{Subdialectal differences in Sorani Kurdish}. In
  \bibinfo{booktitle}{\emph{Proceedings of the third workshop on NLP for
  similar languages, varieties and dialects (vardial3)}}.
  \bibinfo{pages}{89--96}.
\newblock


\bibitem[Marani(2020)]%
        {Marani_2020hac}
\bibfield{author}{\bibinfo{person}{Marani}.} \bibinfo{year}{2020}\natexlab{}.
\newblock \bibinfo{booktitle}{\emph{Elfab\^et\^i Horam\^i}
  (\bibinfo{edition}{1st edition} ed.)}.
\newblock \bibinfo{publisher}{Kurdipedia}.
\newblock
\showISBNx{978-1-55671-216-6}


\bibitem[Maulud et~al\mbox{.}(2023)]%
        {maulud2023towards}
\bibfield{author}{\bibinfo{person}{Dastan Maulud}, \bibinfo{person}{Karwan
  Jacksi}, {and} \bibinfo{person}{Ismael Ali}.}
  \bibinfo{year}{2023}\natexlab{}.
\newblock \showarticletitle{Towards a Complete Kurdish NLP Pipeline: Challenges
  and Opportunities}.
\newblock \bibinfo{journal}{\emph{J. Inform}} \bibinfo{volume}{17},
  \bibinfo{number}{1} (\bibinfo{year}{2023}), \bibinfo{pages}{1--17}.
\newblock


\bibitem[Moseley(2010)]%
        {moseley2010atlas}
\bibfield{author}{\bibinfo{person}{Christopher Moseley}.}
  \bibinfo{year}{2010}\natexlab{}.
\newblock \bibinfo{booktitle}{\emph{Atlas of the World's Languages in Danger}}.
\newblock \bibinfo{publisher}{Unesco}.
\newblock


\bibitem[Mustafa(1971)]%
        {mustafa1971}
\bibfield{author}{\bibinfo{person}{Ezadin Mustafa}.}
  \bibinfo{year}{1971}\natexlab{}.
\newblock \bibinfo{booktitle}{\emph{Serinc\^e le ziman\^i edebi\^i yekgirt\^uy
  kurd\^i}}.
\newblock \bibinfo{publisher}{Salman Al-Azmi Printing House},
  \bibinfo{address}{Baghdad, Iraq}.
\newblock


\bibitem[Nabaz(1976)]%
        {nabaz1976}
\bibfield{author}{\bibinfo{person}{Jamal Nabaz}.}
  \bibinfo{year}{1976}\natexlab{}.
\newblock \bibinfo{booktitle}{\emph{Ziman\^i Yekgirt\^uy Kurd\^i}}.
\newblock \bibinfo{publisher}{National Union of Kurdish Students}.
\newblock


\bibitem[Neamat(2015)]%
        {saber2010derivational}
\bibfield{author}{\bibinfo{person}{Saber Neamat}.}
  \bibinfo{year}{2015}\natexlab{}.
\newblock \bibinfo{booktitle}{\emph{The Derivational Morphology of Word in
  Hawrami and Central Kurdish Dialects}}.
\newblock \bibinfo{publisher}{Salahaddin University}.
\newblock


\bibitem[Paul(2008)]%
        {paul2008kurdish}
\bibfield{author}{\bibinfo{person}{Ludwig Paul}.}
  \bibinfo{year}{2008}\natexlab{}.
\newblock \showarticletitle{Kurdish language I. History of the Kurdish
  language}.
\newblock \bibinfo{journal}{\emph{Encyclopaedia Iranica}}
  (\bibinfo{year}{2008}).
\newblock


\bibitem[Qing-chao et~al\mbox{.}(2021)]%
        {9442500}
\bibfield{author}{\bibinfo{person}{Ni Qing-chao}, \bibinfo{person}{Yin
  Cong-jue}, {and} \bibinfo{person}{Zhao Dong-hua}.}
  \bibinfo{year}{2021}\natexlab{}.
\newblock \showarticletitle{Research on Small Sample Text Classification Based
  on Attribute Extraction and Data Augmentation}. In
  \bibinfo{booktitle}{\emph{2021 IEEE 6th International Conference on Cloud
  Computing and Big Data Analytics (ICCCBDA)}}. \bibinfo{pages}{53--57}.
\newblock
\urldef\tempurl%
\url{https://doi.org/10.1109/ICCCBDA51879.2021.9442500}
\showDOI{\tempurl}


\bibitem[Rashid et~al\mbox{.}(2018)]%
        {rashid2018automatic}
\bibfield{author}{\bibinfo{person}{Tarik~A Rashid}, \bibinfo{person}{Arazo~M
  Mustafa}, {and} \bibinfo{person}{Ari~M Saeed}.}
  \bibinfo{year}{2018}\natexlab{}.
\newblock \showarticletitle{Automatic Kurdish text classification using KDC
  4007 dataset}. In \bibinfo{booktitle}{\emph{Advances in Internetworking, Data
  \& Web Technologies: The 5th International Conference on Emerging
  Internetworking, Data \& Web Technologies (EIDWT-2017)}}. Springer,
  \bibinfo{pages}{187--198}.
\newblock


\bibitem[Rashid et~al\mbox{.}(2017)]%
        {rashid2017systematic}
\bibfield{author}{\bibinfo{person}{Tarik~A Rashid}, \bibinfo{person}{Didar~D
  Rashad}, \bibinfo{person}{Hiwa~M Gaznai}, {and} \bibinfo{person}{Ahmed~S
  Shamsaldin}.} \bibinfo{year}{2017}\natexlab{}.
\newblock \showarticletitle{A Systematic Web Mining Based Approach for
  Forecasting Terrorism}. In \bibinfo{booktitle}{\emph{Modeling, Design and
  Simulation of Systems: 17th Asia Simulation Conference, AsiaSim 2017, Melaka,
  Malaysia, August 27--29, 2017, Proceedings, Part II 17}}. Springer,
  \bibinfo{pages}{284--295}.
\newblock


\bibitem[Rzgar(2024)]%
        {Rzgar_2024}
\bibfield{author}{\bibinfo{person}{Mohammad Rzgar}.}
  \bibinfo{year}{2024}\natexlab{}.
\newblock \bibinfo{title}{Ş\^ewezar\^i Hewram\^i}.
\newblock
\newblock
\urldef\tempurl%
\url{https://bit.ly/3StnVIl}
\showURL{%
\tempurl}


\bibitem[Saeed et~al\mbox{.}(2018)]%
        {saeed2018evaluation}
\bibfield{author}{\bibinfo{person}{Ari~M Saeed}, \bibinfo{person}{Tarik~A
  Rashid}, \bibinfo{person}{Arazo~M Mustafa}, \bibinfo{person}{Rawan A
  Al-Rashid Agha}, \bibinfo{person}{Ahmed~S Shamsaldin}, {and}
  \bibinfo{person}{Nawzad~K Al-Salihi}.} \bibinfo{year}{2018}\natexlab{}.
\newblock \showarticletitle{An evaluation of Reber stemmer with longest match
  stemmer technique in Kurdish Sorani text classification}.
\newblock \bibinfo{journal}{\emph{Iran Journal of Computer Science}}
  \bibinfo{volume}{1} (\bibinfo{year}{2018}), \bibinfo{pages}{99--107}.
\newblock


\bibitem[Salh and Nabi(2023)]%
        {salh2023kurdish}
\bibfield{author}{\bibinfo{person}{Dana~Abubakr Salh} {and}
  \bibinfo{person}{Rebwar~Mala Nabi}.} \bibinfo{year}{2023}\natexlab{}.
\newblock \showarticletitle{Kurdish Fake News Detection Based on Machine
  Learning Approaches}.
\newblock \bibinfo{journal}{\emph{Passer Journal of Basic and Applied
  Sciences}} \bibinfo{volume}{5}, \bibinfo{number}{2} (\bibinfo{year}{2023}),
  \bibinfo{pages}{262--271}.
\newblock


\bibitem[Sami and Hassani(2023)]%
        {sami2023sentiment}
\bibfield{author}{\bibinfo{person}{Maryam Sami} {and} \bibinfo{person}{Hossein
  Hassani}.} \bibinfo{year}{2023}\natexlab{}.
\newblock \showarticletitle{Sentiment Analysis of Opinions about Online
  Education in the Kurdistan Region of Iraq during COVID-19}.
\newblock \bibinfo{journal}{\emph{Qeios}} (\bibinfo{year}{2023}).
\newblock


\bibitem[Taher and Ahmed(2023)]%
        {taher2023correlation}
\bibfield{author}{\bibinfo{person}{Kazheen~Ismael Taher} {and}
  \bibinfo{person}{JIHAN~Abdulazeez Ahmed}.} \bibinfo{year}{2023}\natexlab{}.
\newblock \showarticletitle{Correlation Evaluation Scale Through Text Mining
  Algorithms and Implementation on the Kurdish Language: A Review}.
\newblock \bibinfo{journal}{\emph{QALAAI ZANIST JOURNAL}} \bibinfo{volume}{8},
  \bibinfo{number}{2} (\bibinfo{year}{2023}), \bibinfo{pages}{1312--1337}.
\newblock


\bibitem[Tavadze(2019)]%
        {tavadze2019spreading}
\bibfield{author}{\bibinfo{person}{Givi Tavadze}.}
  \bibinfo{year}{2019}\natexlab{}.
\newblock \showarticletitle{Spreading of the Kurdish language dialects and
  writing systems used in the Middle East}.
\newblock \bibinfo{journal}{\emph{Bull. Georg. Natl. Acad. Sci}}
  \bibinfo{volume}{13}, \bibinfo{number}{1} (\bibinfo{year}{2019}).
\newblock


\bibitem[Zhang et~al\mbox{.}(2022)]%
        {zhang2022can}
\bibfield{author}{\bibinfo{person}{Shiyue Zhang}, \bibinfo{person}{Ben Frey},
  {and} \bibinfo{person}{Mohit Bansal}.} \bibinfo{year}{2022}\natexlab{}.
\newblock \showarticletitle{How can NLP help revitalize endangered languages? a
  case study and roadmap for the Cherokee language}.
\newblock \bibinfo{journal}{\emph{arXiv preprint arXiv:2204.11909}}
  (\bibinfo{year}{2022}).
\newblock


\end{thebibliography}

%%
%% If your work has an appendix, this is the place to put it.
\appendix

% \section{Research Methods}

\end{document}